\documentclass{article}

\usepackage{amsmath,amsfonts,amssymb,times,graphicx,natbib,algorithm,algorithmic,hyperref}

\usepackage[accepted]{data4good2016}

\icmltitlerunning{Predicting Student Dropout in Higher Education}

\begin{document}

\twocolumn[
\icmltitle{Predicting Student Dropout in Higher Education}

\icmlauthor{Lovenoor Aulck}{laulck@uw.edu}
\icmlauthor{Nishant Velagapudi}{nishray@uw.edu}
\icmlauthor{Joshua Blumenstock}{joshblum@uw.edu}
\icmlauthor{Jevin West}{jevinw@uw.edu}
\icmladdress{DataLab, The Information School, University of Washington,
            Seattle, WA 98195, USA}

\vskip 0.3in
]

\begin{abstract}
Each year, roughly 30\% of first-year students at US baccalaureate institutions do not return for their second year and over \$9 billion is spent educating these students. Yet, little quantitative research has analyzed the causes and possible remedies for student attrition. Here, we describe initial efforts to model student dropout using the largest known dataset on higher education attrition, which tracks over 32,500 students' demographics and transcript records at one of the nation's largest public universities. Our results highlight several early indicators of student attrition and show that dropout can be accurately predicted even when predictions are based on a single term of academic transcript data. These results highlight the potential for machine learning to have an impact on student retention and success while pointing to several promising directions for future work.
\end{abstract}

\section{Introduction}
\label{sec:intro}

Student dropout is a major concern in the education and policy-making communities \cite{demetriou2011integration,tinto2006research}. About 40\% of students seeking bachelor's degrees do not complete their degree within 6 years \cite{national_center_for_education_statistics_nces_????} with universities losing tens of billions of dollars in revenue each year \cite{raisman2013cost}. First-year student attrition is of particular importance, as United States (US) state and federal governments spent over \$9 billion from 2003-2008 on educating the 30\% of full-time first-year students seeking baccalaureate degrees who do not return for a second year \cite{schneider2010finishing}.

Much of the groundwork for theories on student post-secondary attrition was laid in the 1970s-1980s with the work of Tinto \yrcite{tinto1975dropout,tinto1987leaving}, Spady \yrcite{spady1970dropouts}, and Bean \yrcite{bean1980dropouts}, to name a few. Despite long-standing theory, student drop out continues to be a large concern to the education community and policy makers as attriting students lose time and effort in their failed pursuits while institutions have no recourse to recoup the scarce resources they devoted to the students. 
In recent years, the rise of massive online open courses (MOOCs) and other online educational environments has seen an increase in the application of data mining and machine learning techniques to educational data, particularly in the domains of educational data mining and learning analytics \cite{baker2009state,siemens2011penetrating,siemens2012learning,romero2013data, baker2014educational}. However, in part due to the lack of appropriate data, much less quantitative research has focused on student dropout in the traditional classroom environment.

Here, we model student dropout using data gathered from the registrar databases of a large, publicly-funded, four-year university in the US. To our knowledge, this is the largest dataset used to study student attrition at scale. Our broader objective is to understand the key determinants of dropout, to accurately identify students likely to attrite, and to recommend policy interventions to reduce student attrition. In this workshop paper, we focus on results from our initial attempts to predict student dropout using demographic information as well as transcript records from the student's first academic term at the university.

Our work relates to recent efforts to analyze student dropout in small, homogeneous populations. For instance, Dekker et al. use a host of machine learning approaches to predict student drop out among a group of 648 students in the Electrical Engineering department at the Eindhoven University of Technology using the first semester's grades \yrcite{dekker2009predicting}. Kova{\u c}i{\'c} use tree-based learning methods while focusing on feature selection to conduct a similar analysis within the Open Polytechnic of New Zealand, relying on socio-demographic features of 453 students \yrcite{kovavcic2010early}. Moseley and Mead, meanwhile, use rule induction to predict dropout at the course level among 528 nursing students \yrcite{moseley2008predicting}. Lin et al. rely on neural networks to model retention in engineering with 1,508 students \yrcite{lin2009student}. Most closely related to our work, Delen and Thammasiri use machine learning techniques to predict whether freshmen will enroll for a second term (not their eventual graduation) \cite{delen2011predicting,thammasiri2014critical}. While promising, most of these previous studies have focused on subsets of very homogeneous students in particular fields of study. Our approach, by contrast, considers an extremely heterogeneous population at one of the nation's largest public universities.

\section{Methods}

\subsection{Data}
De-identified, psuedonymized data were gathered from the University of Washington's (UW) registrar databases in the summer of 2013. The data contain the demographic information (race, gender, birth date, resident status, and identification as Hispanic), pre-college entry information (SAT and ACT scores, if available), and complete transcript records (classes taken, time at which they were taken, grades received, and majors declared) for all students in the University of Washington (UW) system (consisting of a main campus at Seattle and two satellite campuses: Bothell and Tacoma). Our focus is on matriculated undergraduate students at the main campus who first enrolled between 1998 and 2006. The year 2006 was used as an upper bound to allow for 6 full years to graduate from the time of first enrollment. In all, this was 69,116 students. Of these 69,116, 5 students did not have birth years available and were consequently excluded from the analysis. The overall graduation rate in this dataset was about 76.5\% based on the definition of non-completion (NC) presented in Section \ref{sec:NCs}. We randomly sampled from the majority class to create a balanced dataset consisting of 32,538 students for this preliminary work. In all, about half the data was comprised of freshmen entrants, while transfers from 2-year colleges and transfers from 4-year colleges each comprised about one-quarter of the dataset. An overview of the demographics and graduation rates across those demographic groups are presented in Table~\ref{table:demTable}.

\begin{table}[h]
\caption{Demographics in (balanced) dataset}
\label{table:demTable}
\vskip 0.15in
\begin{center}
\begin{small}
\begin{sc}
\begin{tabular}{lcc|c}
\hline
\abovespace\belowspace
 & Grads & NCs & Grad rate \\
\hline 

\abovespace\belowspace
All    & 16,269 & 16,269 & 50.00\% \\

\hline
\abovespace
\hspace{0.3cm} \textit{Gender}  & & & \\
Females & 8,790 & 8,134 & 51.94\% \\
Males    & 7,462 & 8,129 & 47.86\% \\
\belowspace
Other/Unknown  &  17 & 6 & 73.91\% \\

\hline
\abovespace
\hspace{0.3cm}\textit{Previous Schooling} & & & \\
Freshmen & 8,685 & 7,488 & 53.70\% \\
Trans. from 2-yr    & 4,125 & 4,162 & 49.78\% \\
\belowspace
Trans. from 4-yr  &  3,459 & 4,619 & 42.82\% \\

\hline
\abovespace
\hspace{0.3cm}\textit{Race/Ethnicity} & & & \\
Afr. American     & 439 & 642 & 40.61\% \\
Amer. Indian      & 196 & 322 & 37.84\% \\
Asian      & 3,843 & 3,571 & 51.83\% \\
Caucasian  & 9,317 & 9,155 & 50.44\% \\
Hawaiian/Pac. Is. & 89 & 142 & 38.53\% \\
\belowspace
Other/Unknown & 2,385 & 2,437 & 49.46\% \\

\hline
\abovespace
\hspace{0.3cm}\textit{Hispanic} & & & \\
Hispanic & 657 & 821 & 44.45\% \\
\belowspace
Not Hispanic & 15,612 & 15,448 & 50.26\% \\

\hline
\abovespace
\hspace{0.3cm}\textit{Residency Status} & & & \\
Resident & 14,533 & 14,116 & 50.73\% \\
\belowspace
Non-resident & 1,736 & 2,153 & 44.64\%\\

\hline
\end{tabular}
\end{sc}
\end{small}
\end{center}
\end{table}

\subsection{Defining Non-Completion}
\label{sec:NCs}
Students who dropped out (non-completions or NCs) are defined as those students who did not complete at least one undergraduate degree within 6 calendar years of first enrollment. In the dataset, this consisted of a single, binary outcome feature. The UW uses a quarter-based system of enrollment and this 6-year time to completion translated to 24 calendar quarters after the quarter of first enrollment. Enrollment in this case was defined as when a student received at least one transcript grade (regardless of whether it is numeric or passing) for a term. Transfer students' time in the university system was accounted for by dividing the number of credits transferred to the university by the number of credits needed per quarter to graduate in 12 quarters (i.e. 4 years without taking summer classes).

\subsection{Feature Mapping}
\label{sec:features}
Race, gender, and resident status were categorical variables where each student only belonged to a single category and the inclusion in categories was mutually exclusive across variables. Each possible race (6 total), gender (3 total), and resident status (7 total but grouped in Table 1) were mapped across dummy variables. SAT and ACT scores, meanwhile, were only available for 40\% and 12\% of the data, respectively. To impute missing SAT and ACT scores, we used a linear regression model with other demographic and pre-college entry data. We also used mean imputation for these missing values and obtained results similar to those presented.

The top 150 most frequently declared majors during the first term were similarly mapped across dummy variables. Each department in which students took classes was mapped across four features for each student: a binary variable indicating whether the student took a class in that department, a count of the number of credits taken in that department by the student, a count of the number of classes taken in that department by the student, and the grade point average (GPA) of the student for all graded classes taken in that department. This resulted in 784 additional features. The same four categories of features were also used to look at ``gatekeeper'' classes in science, technology, engineering, and math (STEM) fields, namely: entry-level physics, chemistry, biology, and math classes, which are typically taken as year-long sequences. Indicators of whether the student took remedial classes during the first term or were part of a first-year interest group (FIG)\footnote{For more information, see \href{http://fyp.washington.edu/first-year-interest-groups/}{here}} were also included. It should be noted that we did not have any information regarding students' financial standing or history, though we understand this to be a large part of the motivation behind students' decisions to stop their studies \cite{cabrera1992role}.

\subsection{Experiments}

Below, we report results from three machine learning models (regularized logistic regression, k-nearest neighbors, and random forests) to predict the binary dropout variable on the features described in Section \ref{sec:features}. In all experiments, we report measures of performance on a 30\% random sample of test data, which is not used in cross-validation or model selection. With the remaining 70\% of the data, we use 10-fold cross-validation to tune the model parameters (e.g. the regularization strength for logistic regression, the number of neighbors in kNN, and the depth of the tree in random forests).

We are also interested in understanding which elements in a student's data are the best predictors of dropout. For this, we run $k$ separate logistic regressions of dropout on the $k^{th}$ feature, trained on the training set, and with performance calculated on the test data. 
Finally, we use a regularized linear regression to predict the number of terms each non-completing student enrolled in before dropping out. K-fold cross validation was used to determine the regularization strength of the model. As before, we randomly sample 70\% of the data (in this case, all non-completions) to tune the regularization parameter and performance is reported on the remaining 30\% of the data.

\section{Results}
\subsection{Predicting dropout}

\begin{figure}[!h]
\vskip 0.2in
\begin{center}
\centerline{\includegraphics[width=\columnwidth, trim = 0 0.5cm 0 0]{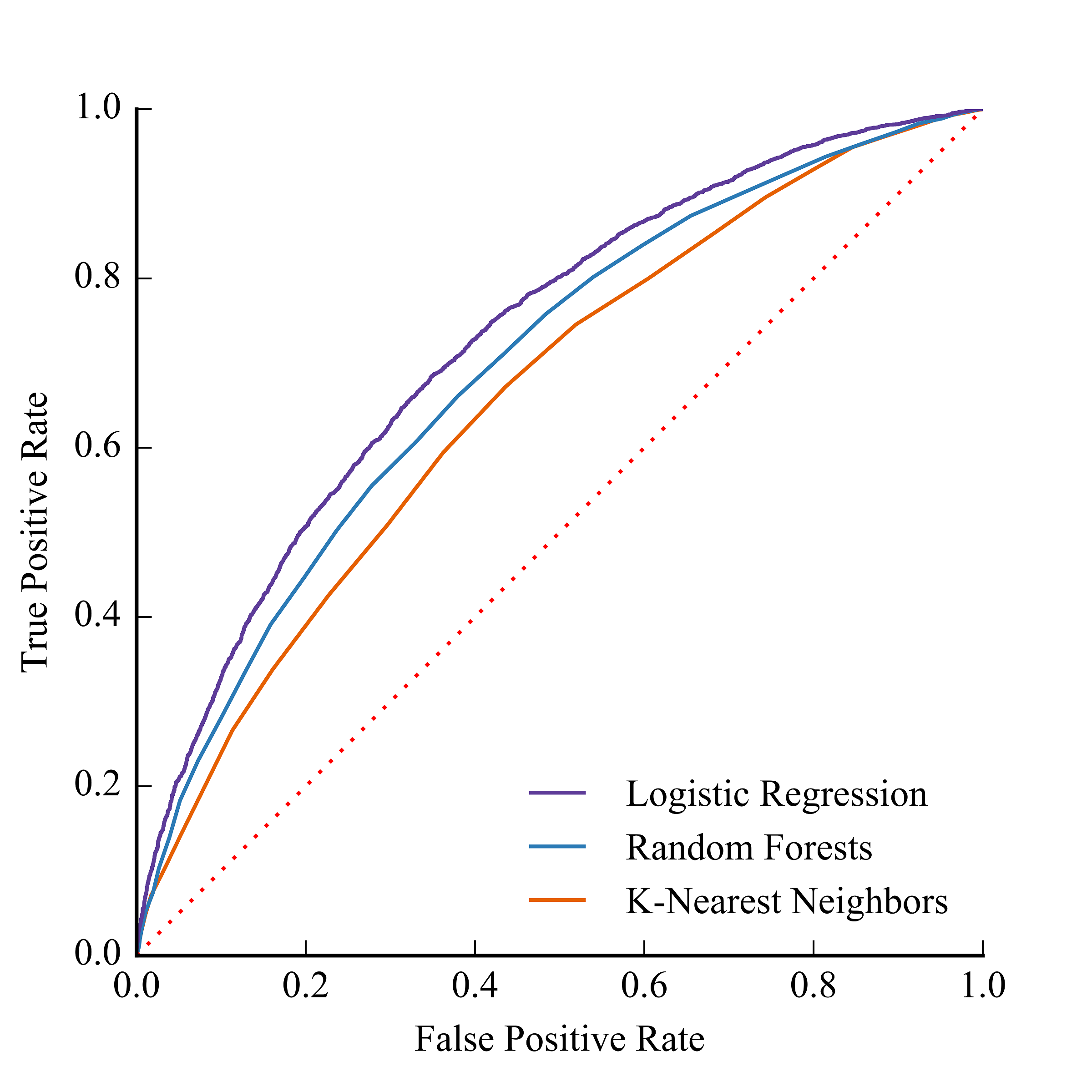}}
\caption{ROC curves}
\label{fig:ROC}
\end{center}
\vskip -0.2in
\end{figure}

Receiver operating characteristic (ROC) curves for each of the three models are shown in Figure \ref{fig:ROC}. The prediction accuracies and AUC for the ROC curves are shown in Table \ref{table:ROCtable}. Performance was comparable across models (AUC between 0.66 and 0.73), with modest gains in the case of logistic regression. As we discuss below, while we believe these initial results indicate a strong signal in the transcript data for predicting student dropout, we view them as relatively naive baselines to be improved upon in future work. In particular, we are actively exploring alternative approaches to the learning task (e.g., neural and spectral methods).

\begin{table}[h]
\caption{Prediction Accuracy and ROC for Models}
\label{table:ROCtable}
\vskip 0.15in
\begin{center}
\begin{small}
\begin{sc}
\begin{tabular}{lcc}
\hline
\abovespace\belowspace
Model & Accuracy & AUC\\
\hline 
\abovespace
Logistic Regression    & 66.59\% & 0.729\\
Random Forests    & 62.24\% & 0.694 \\
\belowspace
K-Nearest Neighbors    & 64.60\% & 0.660\\
\hline
\end{tabular}
\end{sc}
\end{small}
\end{center}
\end{table}

\subsection{Correlates of Dropout}

Examining individual features reveals several interesting trends. Firstly, GPA in math, English, chemistry, and psychology classes were among the strongest individual predictors of student retention. From preliminary analysis of our data, we know that baccalaureate transfers tend to graduate at much lower rates than their peers in our dataset \cite{aulck_using_2016}. As such, indicators of students' previous schooling were also among those with the highest performance. Interestingly, birth year was also a strong predictor of eventual attrition as was the year of first enrollment. This is reflective of the fact that our data exhibits a declining overall trend in attrition across time, with overall attrition rates at 27.6\% for the 1998 entering class and 20.2\% for the 2006 entering class. Interestingly, the first quarter in which a student enrolled (e.g. Autumn, Winter, Spring, or Summer) was also a strong predictor of attrition. In all, however, no single feature yielded a predictive accuracy higher than 54\%. The 7 features with the highest predictive performance in isolation are shown in Table \ref{table:indFeatures}.

\begin{table}[h]
\caption{Features with Highest Predictive Performance}
\label{table:indFeatures}
\vskip 0.15in
\begin{center}
\begin{small}
\begin{sc}
\begin{tabular}{lcc}
\hline
\abovespace\belowspace
Feature & Accuracy & AUC\\
\hline 

\abovespace
GPA in Math Classes    & 52.95\% & 0.571\\
GPA in English Classes    & 53.00\% & 0.567\\
First Qtr. of Enrollment  & 53.49\% & 0.549 \\
GPA in Chemistry Classes    & 51.79\% & 0.549 \\
First Year of Enrollment    & 53.58\% & 0.547 \\
Birth Year    & 53.49\% & 0.545 \\
\belowspace
GPA in Psychology Classes    & 53.49\% & 0.541 \\

\hline
\end{tabular}
\end{sc}
\end{small}
\end{center}
\end{table}

\subsection{Timing of Dropout}

Finally, we were sensitive to the fact that university administrators are eager to know not just \textit{who} is likely to drop-out or \textit{factors} likely to influence student attrition, but also \textit{when} at-risk students are most likely to attrite. Thus, in addition to predicting student attrition, we also sought to predict when a non-completion would eventually stop pursuing their degree using a single term's data. This yielded marginally successful results. We obtained a root mean squared error (RMSE) value of 5.03 quarters when using data on all non-completions. When excluding the bottom 5\% and 10\% of least accurate predictions, RMSE was 4.14 and 3.74 quarters, respectively. On average, non-completions enrolled in 7.35 (SD: $\pm$ 5.65) quarters before stopping their studies.

\section{Future Directions}

The early-stage results described above point to several promising directions for future work. First, we are extending our analysis to the full dataset and to other universities, including dealing with issues related to class imbalances \cite{thammasiri2014critical}. We also intend to expand our analysis beyond the first term's transcript data and take a more comprehensive look at the processes of attrition. Doing so will allow us to better understand the nuances associated with dropout across different disciplines, thereby providing some theoretical context to early warning signs of attrition. As an extension to this, we hope to leverage this work for possible interventions at the policy level to reduce attrition. We are in discussions with administrators at the University of Washington to better interpret our results and pinpointing possible intervening policies. 

We are also interested in technical improvements to our prediction models. As shown in this work, our three initial prediction models yielded similar results. In hopes of improving these models, we intend to first use feature engineering on our existing feature space. Feature engineering is often used in the analysis of customer churn from businesses, as shown recently by Huang et al., for example \yrcite{huang2015telco}, and we see a direct analog with student attrition. We also intend to eventually look at convolutional and recurrent neural networks, thereby reducing the need for hand-engineered features.

% This work focuses on the issue of student attrition from a university-centric perspective. The students may have continued to pursue their education elsewhere and without more comprehensive data linked across institutions, this would be impossible to ascertain. As one way of addressing this constraint, we are speaking with other universities that are willing to compare their data and results to ours.  

\section{Conclusions}

In this work, we show preliminary results for predicting student attrition from a large, heterogeneous dataset of student demographics and transcript records. Predicting eventual student attrition yielded promising results from a balanced dataset of over 32,500 students with regularized logistic regression providing the strongest predictions. GPA in math, English, chemistry, and psychology courses were among the strongest individual predictors of attrition, as were year of enrollment and birth year, thus highlighting time effects in our data. Predicting the number of quarters non-completions take prior to dropping out yielded marginal results, as predictions had an RMSE of about 5 quarters of enrollment. Next steps will involve discussions with university administrators, improving our predictive models, and possibly even expanding our dataset to other universities and community colleges where attrition rates tend to be much higher.

\newpage{}
\newpage{}
\bibliography{icml_bib}
\bibliographystyle{icml2016}

\end{document}